%% file: main.tex
\documentclass[twocolumn]{article}

\usepackage[width=17cm,height=22cm]{geometry}

\usepackage[utf8]{inputenc}
\usepackage[english]{babel}
\usepackage{fancyvrb}
\usepackage{authblk}
\usepackage{hyperref}
\usepackage{graphicx}
\usepackage{xcolor}
\definecolor{bluep}{rgb}{0.2, 0.3, 0.6}
\usepackage{enumitem}

\title{Baselines and a datasheet for the Cerema AWP dataset}
\author[2,3]{Ismaïla Seck}
\author[1]{Khouloud Dahmane}
\author[1]{Pierre Duthon}
\author[2]{Gaëlle Loosli}
\affil[1]{Cerema, Clermont-Ferrand}
\affil[2]{Université Clermont Auvergne, LIMOS}
\affil[3]{INSA de Rouen, LITIS}
\date{}
\begin{document}
\maketitle

\begin{abstract}
 This paper presents the recently published Cerema AWP (Adverse Weather Pedestrian) dataset for various machine learning tasks and its exports in machine learning friendly format. We explain why this dataset can be interesting (mainly because it is a greatly controlled and fully annotated image dataset) and present baseline results for various tasks. Moreover, we decided to follow the very recent suggestions of {\em datasheets for dataset}, trying to standardize all the available information of the dataset, with a transparency objective.
\end{abstract}

\medskip

\noindent\textbf{Mots-clef}: Datasheet for dataset, Cerema AWP, classification, regression, auto-encoder, GAN.

\section{Introduction}
\label{sec:intro}
Deep learning techniques have shown impressive results during the past years, in particular when dealing with image processing. However there remains a lot of open questions, for instance in terms of quality evaluation and method comparisons.

The introduction of new datasets
and of new algorithms leads to new baselines. That raises some questions about how those baselines are established, about the fairness in the comparison between the newly introduced algorithm and the ones previously known.This problem is not new \cite{armstrong_relative_2009} but it may be increased by the huge amount of publications in the field nowadays. Let us take the case of Generative Adversarial Networks  \cite{goodfellow_generative_2014}, there are many variants of GAN\footnote{ cf https://github.com/hindupuravinash/the-gan-zoo}, often with the claim to outperform previous versions and so with no objective metric to back that claim.\footnote{ more detail about baselines and objective evaluation for GAN threadreaderapp.com/thread/978339478560415744.html}

Another aspect linked to new datasets is the need for transparency and good understanding of its content. The recent work-in-progress \cite{gebru_datasheets_2018} discusses the need for standardization of datasets we use. While the cited paper is only a work-in-progress proposal to a standard datasheet content that will likely change in the coming months/years, we decided to make an attempt of description of the {\em Cerema AWP dataset} according to this proposal. 

\begin{figure}[ht]
\includegraphics[width=0.49\linewidth]{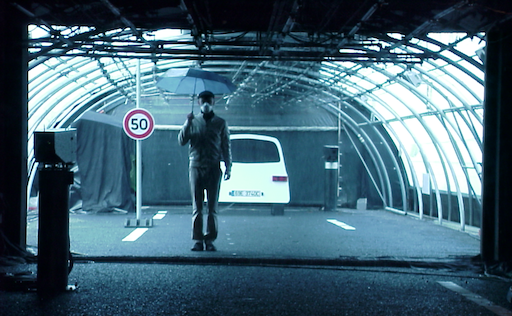}
\includegraphics[width=0.49\linewidth]{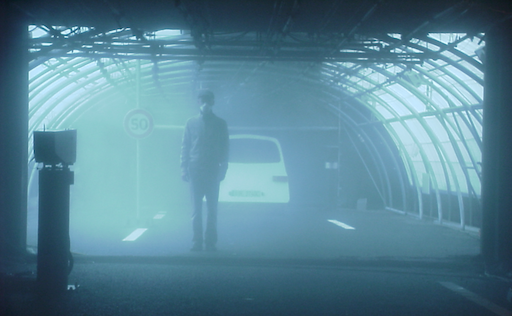}
\includegraphics[width=0.49\linewidth]{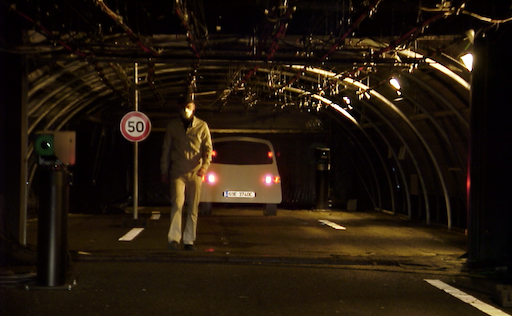}
\includegraphics[width=0.49\linewidth]{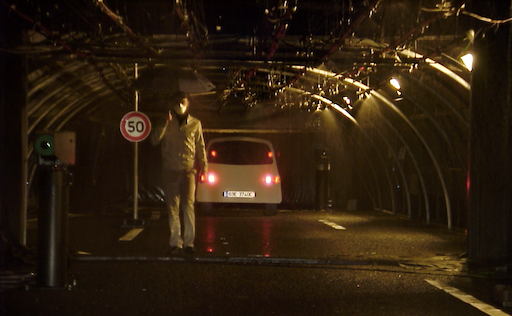}
\caption{\label{fig:cerema} Examples of the Cerema AWP dataset. The same pedestrian appears roughly at the same place in different weather and light conditions. }
\end{figure}

The Cerema AWP (Adverse Weather Pedestrian) dataset \cite{dahmane_cerema_2017}\cite{duthon_descripteurs_2017} was released recently\footnote{\url{https://ceremadlcfmds.wixsite.com/cerema-databases}}. It is a very particular annotated image dataset since all images are obtained from a fully controlled environment. Each image shows exactly the same scene, as illustrated on figure \ref{fig:cerema}, but under various controlled weather conditions.

The interest of such a dataset relies on the precise knowledge of its content. Of course it might not be useful for real life machine learning applications, but we believe that it could help in the evaluation of machine learning methods, such as controlled image generation. 

\section{Cerema AWP datasheet}
\label{sec:cerema}

The idea of creating datasheets for datasets \cite{gebru_datasheets_2018} comes from several observations of major failures of machine learning systems due to biases or internal flaws in the dataset that were used to train them. Most of times, the misuse of datasets are unintentional and due to the fact that very few factual information is known about publicly available datasets. Such a document, with standard questions, would have at least two targets : on the one hand, the {\em consumers} of the data, who would be informed of how and what for were the data collected, and on the other hand, the {\em providers} who would have the opportunity to improve their dataset quality and maybe focus on point that would have been ignored otherwise (such as legal aspects or fairness). 

We present the datasheet for Cerema AWP, as it is suggested in \cite{gebru_datasheets_2018} as an attempt to follow their guidelines and hopefully participate to this very interesting initiative aiming at bringing more fairness in machine learning. The datasheet is organized exactly as proposed and is placed at the end of the paper on page \pageref{sec:datasheet} for readability issues. 

To sum-up the content of the dataset, it was originally designed to train more robust pedestrian detectors, starting from the observation that most of the existing datasets were containing favorable weather conditions. In a special installation ({\em the Cerema Fog and Rain R\&D platform} ), a tunnel in which rain, fog and night can be artificially created, it was possible to take photos of 5 persons walking back and forth in the tunnel, wearing different clothes and under different weather conditions. Using classical detectors on those images, it was observed that indeed they were failing much more under adversarial weather conditions. 

In this paper we are not interested directly on the pedestrian detection but we rather want to exhibit other interests of this particular dataset.

\section{Tasks and baseline results}
\label{sec:tasks}

In this section, we present several identified tasks for the Cerema AWP and some baseline results. All presented baselines are convolutional neural networks, which are well adapted to image processing and large datasets. For each proposed model, we tried to produce the best models we could, using only standards layers. 
In the following, we first present how we pre-processed and split the dataset.

\input{tasks}

\section{Conclusion}
\label{sec:conlusion}
To conclude this paper, we first summarize our baseline results for the Cerema AWP dataset. It is very likely that these score can be improved with more elaborated models or having the chance to run experiments on a bunch of GPU to practice a better model selection. However we think that establishing and publishing a fair baseline before applying fancy methods on this new dataset will produce higher quality results in time. 

We summarize in table \ref{tab:test} the results on the test set for each supervised task. Those results were obtained once, for our best models selected on validation results.
\begin{table}[ht]
\centering \small
\begin{tabular}{lcc}
\hline
Task & Accuracy & Acc in MultiLabel\\
\hline \hline
Weather & 99.5\% & 99.9\%-100\%\\
Pedestrian & 100\%& 100\%-100\%\\
Clothes & 99.6\%  & 99.0\%-99.1\%\\
Direction & 81.4\% & 82.1\%-83.6\%\\
PedestrianClothes & 99.5\%& - \\
\hline
Reg on Boxes & MSE : 0.04 &  MAE : 0.003\\
\hline
\end{tabular}
\caption{\label{tab:test} Final results of our baseline, on the test set. For results in the multi label setting, the first value in for unweighted costs and the second for the weighted costs.}
\end{table}

The purpose of this paper was two fold : 
\begin{itemize}
\item presenting a dataset that we believe can be useful for some systematic exploration of some algorithms properties, in an attempt of standardization, 
\item and establishing baseline results before any advanced exploitation, in order to reduce the bias in later results presentation.
\end{itemize}
This work was greatly influenced by current discussions inside and outside the community about fairness, results quality and comparability. While we know that publishing datasheets and baselines won't solve many issues, we believe that is at least a step in the good direction. 

Our future work concerning this database is quite evidently to take advantage of its particularity to push generative models towards unexplored behaviors and to work on the quality evaluation of produced images, using a validation dataset, which is mainly impossible with existing standards dataset.

\subsection*{Acknowledgment}
This work has been partly supported by the grant ANR-16-CE23-0006 “Deep in France” and by the French National Research Agency within the "Investissement d'avenir" program via the LabEx IMobS3 (ANR-IO-LABX-16-01). 

\scriptsize
\bibliography{Zotero}

\input{datasheet}

\input{appendix}

\end{document}

%% file: tasks.tex
\subsection{Pre-processing}
The dataset provided on the original website consists of raw images stored in a hierarchy of folders, which gives some of the labels:
\begin{itemize}[noitemsep,topsep=0pt,parsep=0pt,partopsep=0pt]
	\item Level 1 : Weather conditions (5 with daylight, labels beginning with 'J'), 5 during night (labels beginning with 'N'). The 5 are good weather (labels 'JCN' and 'NCN'), low rain ('JP1','NP1'), heavy rain ('JP2','NP2'), little fog ('JB1','NB1'), heavy fog ('JB2','NB2')).
    \item Level 2 : Pedestrian (5 different persons)
    \item Level 3 : Clothes (2 different jackets colors)
\end{itemize}
The other labels and information (such as the pedestrian direction, the bounding box or the exposure time) are provided in a separated csv file. This file also gives a redundant information concerning pedestrian and clothes, in the form of a new label that separates the dataset into 10 classes (combination of pedestrian identification and clothes). 

We decided to export all the information (images and labels) in a single format in order to simplify the information management.
The proposed script\footnote{available on github \url{https://github.com/gloosli/ceremaAWP}} also provides an option for image downsizing.
Here is a final format of the hdf5 file for Python: 
\begin{description}[align=right,labelwidth=3.5cm,noitemsep,topsep=0pt,parsep=0pt,partopsep=0pt]
	\item [attribute 'width'] : width of stored images
	\item [attribute 'height'] : height of stored images
    \item [dataset 'train'] : stores 85\% of the dataset
    \item [dataset 'test'] : stores the remaining 15\% of the dataset
\end{description}
The split between train and test is done such that both sets have similar proportion of each class (10 classes weather, 10 classes pedestrian, 6 classes direction). On average, the test set contains about 12 example of each  of the 600 combinations of classes.
The information inside train and test is stored as follows:
\begin{description}[align=right,labelwidth=3.5cm,noitemsep,topsep=0pt,parsep=0pt,partopsep=0pt]
	\item[weather] : list of labels from 0 to 9 (order compared to the original dataset : ['JB1', 'JB2', 'JCN', 'JP1', 'JP2', 'NB1', 'NB2', 'NCN', 'NP1', 'NP2'])
  	\item[pedestrian] : list of labels from 0 to 4
   	\item[clothes] : list of labels from 0 to 1
   	\item[pedestrianClothes] : list of labels from 0 to 9 (combination of the two previous ones)
 	\item[direction] : list of labels from 0 to 5
  	\item[exposition] : list of  reals
  	\item[boundingBox] : list of size (nbElements, 4), [x,y,width,height] in proportion of height and width (originally this information is given in pixels, but since we propose to extract images at different scales of downsizing, we preferred proportions)
   	\item[images] : Numpy matrix (nbElements $\times$width$ \times $height$\times$3)
\end{description}

Unless mentioned otherwise, results are provided on images of size 1/4 of the original size (256x158). For all experiments, the train set is split randomly in order to have results on a 20\% validation set. 

\subsection{Classification tasks}
We applied standard convolutional networks, using Keras on Tensorflow, to each of the classification tasks. The summary of the architectures can be found in the annexe, the optimization algorithm is Adam with defaults parameters, except for the learning rate set at $10^-5$.
We expect the results to be very high on most of the classes: due to the large similarity between images, we can foresee that some sort of over-fitting will do well. 

\paragraph{Easy classification tasks}
We summarize here results for Weather, Pedestrian, Clothes, and PedestrianClothes, since each of them can be easily solved with validation performance higher than 99\%. From the simplest (the weather, since the relevant information is mainly contained in the color range) for which 1 epoch is enough, to the least simple (PedestrianClothes) which requires 10 epochs, we observe that there is no real challenge. 

\paragraph{Classification on the Direction labels}
The direction label indicates how the pedestrian is shown on the image : face, back, left, right (close or far), diagonal from far right to close left. This classification task is less obvious since the algorithm has to focus on details on the pedestrian only. However it is possible that it can infer results from the position in the image because of the regularity of the path. 
After 10 epochs, the accuracy on the validation set is quite stable and stays at around 80\%.
We observe the confusion matrix (table \ref{tab:confusionMatrix}) and see that some distinctions are not easy (like between 3 and 4 or between 5 and 6), which make sense since the positions in the image are similar for these couple of classes, but also between 1 and 2, which is more surprising. On the contrary, directions 1 and 2 are almost never mismatched with the other directions. From this we conclude that indeed the position in the image is used to detect the direction, but not only (5 and 6 are exactly at the same locations but are rather well distinguished overall). 

\begin{table}[h]
\scriptsize\centering 
\begin{tabular}{|c|c|c|c|c|c|c|}
\hline
true/pred & 1 & 2 & 3 & 4 & 5& 6\\
\hline
 1 & \bf 726 & 197 &    0 &    0&    24 &    5\\
\hline
 2 &145&  \bf  1382&   72&    0&   54&   75\\
\hline
 3&   8&  155&   \bf 754&  118&   11&    9\\
\hline
 4&  7&   38&  163&   \bf 487&   43&    1\\
\hline
 5&   6&    9&    2&   28&  \bf 1727&   57\\
\hline
 6&   0&   14&    0&    4&  182&  \bf  1193\\
\hline
\end{tabular}
\caption{\label{tab:confusionMatrix} Confusion matrix for the direction detection}
\end{table}

\subsection{Multi-label classification}
We decided for this experiment to build an architecture that will share the convolutional part but have separate output for each class, in order to obtain multi-label classification. Labels that are taken here are {\em weather, pedestrian, clothes, direction}. For pedestrian and clothes, we decided to add a common layer as shown in the annexe. Based on the individual performance and number of required epochs of each classification tasks, the losses are  weighted as follows : 
\begin{itemize}[noitemsep,topsep=0pt,parsep=0pt,partopsep=0pt]
\item weather : 0.2
\item pedestrian : 0.2
\item clothes : 0.5
\item direction : 1
\end{itemize}
We also trained the model without weighting losses to compare.

\begin{table}[h!]
\centering \small
\begin{tabular}{lcc}
\hline
& Without weights & With weights \\
\hline \hline
Acc Weather & 100\% & 100\%\\
Acc Ped & 100\% & 100\%\\
Acc Clothes &  99.3\% & 99.2\%\\
Acc Direction & 82.1\% & 85.0\%\\
\hline
\end{tabular}
\caption{\label{tab:multilab} Results for different loss weights after 10 epochs}
\end{table}

From table \ref{tab:multilab}, we observe that easy classes are perfectly detected in this setting, but also that the Direction detection seems to be improved. We suspect that sharing the feature extraction part with tasks that are directly related to the pedestrian characteristics (such as clothes) helps the model to focus on the interesting part of the image, which is the pedestrian. We also observe that the loss weighting has a positive impact on the Direction detection.  

\subsection{Regression}
The bounding box information comes as a vector of size 4, containing (x,y) of the top left corner and (w,h), the width and height of the box. All values are between 0 and 1 and represent a percentage of image width and height. For this experiment we treat those vectors as a multi-regressor problem. The architecture is the similar to the classification tasks, with an output shape of 4. We achieve a mean square error (MSE) on the validation set of 0.0020 and a mean absolute error (MAE) of 0.0327.
\begin{figure}[htb]
\includegraphics[width=0.49\linewidth]{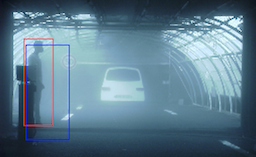}
\includegraphics[width=0.49\linewidth]{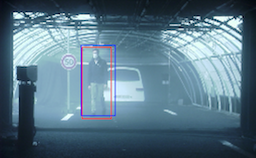}

\includegraphics[width=0.49\linewidth]{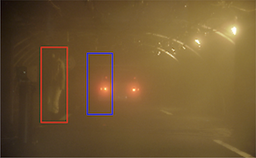}
\includegraphics[width=0.49\linewidth]{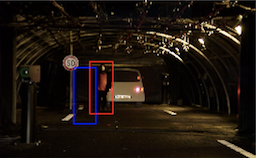}
\caption{\label{fig:reg} Examples of good and bad predicted bounding boxes. The red box is the true one, the blue box is the predicted one.}
\end{figure}
From those results it's hard to conclude on the interest of this tasks using only regression : it's not obvious that the machine actually learns to detect the pedestrian. In order to have a more precise idea, we decided to make another experiment in which we train on a selected subsample of the dataset, such that pedestrian in the validation set will appear at places that were not presented in the train set. To achieve this, we use the direction information and simulate a covariate shift problem \cite{quionero-candela_dataset_2009}. Indeed, as illustrated on figure\ref{fig:tunnel}, the directions are linked to the positions. We decided to exclude from the training set all images with directions 1 and 2 (labels 0 and 1). 
For the evaluation, we present only the images with directions 1 and 2, or images with any directions. This experiment confirms that the method can actually detect a pedestrian at an unusual place but with twice the amount of errors. 

\begin{table}[htb]
\centering \small
\begin{tabular}{ccc}
\hline
 & Excluded directions & All directions \\
 \hline 
 \hline
 MSE & 0.0620 & 0.0393 \\
 MAE & 0.0066 & 0.0031 \\
\hline
\end{tabular}
\caption{\label{tab:reg2} Results of the regression tasks when training with a subpart of the dataset, such that some pedestrian position do not appear in the training set. The two tests sets are composed by either only images of the remaining positions, or all images. We see from those results that the proposed method actually uses the position in the image to detect the pedestrian}
\end{table}

\subsection{Image reconstruction and generation}
\subsubsection{Auto-encoders}
\begin{figure}[h!]
\includegraphics[width=\linewidth]{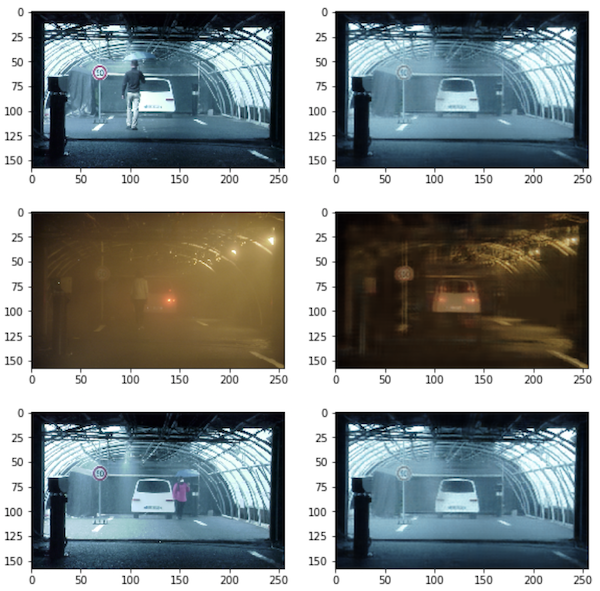}
\caption{\label{fig:AE} Examples of reconstructed images using auto-encoders. On the left columns are true images, on the right column are the generated images. We observe that no pedestrian is generated, only the recurrent background is reconstructed.}
\end{figure}

From now on, we focus on image reconstruction and generation, starting with auto-encoders \cite{hinton_autoencoders_1994}. The idea of autoencoders is to train a neural network that learns to reproduce its input as output, with a smaller size intermediate layer. The representation inside the intermediate layer is a compression of the input. Autoencoders can be staked to obtain successive compressions of the input. 
Here we use autoencoders to evaluate how difficult it is for a neural network to generate our images. Figure \ref{fig:AE} shows the obtained results, based on convolutional layers. We see that it is quite easy to make the autoencoder catch the information on the background and weather conditions. However, we were not able to make it print the pedestrians. 

\subsubsection{GAN}

Generative adversarials Networks,GANs, were introduced in 2014 \cite{goodfellow_generative_2014} and since then dragged the interest of many. The idea is to train simultaneously two networks the generator $G$ and the discriminator $D$. Working for example with an image dataset, the aim of the discriminator $D$ is predict whether or not an image presented to it is from the dataset or from the generator $G$. This simultaneous training of those networks can be seen as a two-player minimax game which has a unique solution. At that solution, neither the discriminator nor the generator can improve anymore. Hopefully, by the time the training get that optimal point, images generated can not be told apart by the discriminator, and that stands for humans too. Since the introduction of GANs many versions have been introduced but \cite{lucic_are_2017} states that all those variants are at some points equivalent, we use a DCGAN-like \cite{radford_unsupervised_2015} which is suitable in our experiments. It is well known that GANs are difficult to train, and this difficulty increase with the variability of the input images, and also their size. 

\begin{figure}[htb]
\includegraphics[width=\linewidth]{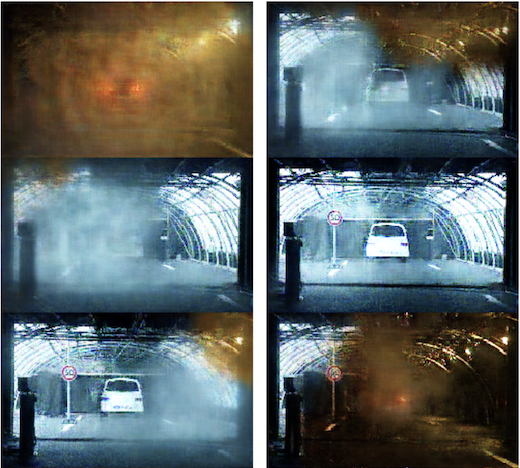}
\caption{\label{fig:GAN} Examples of generated images using a DCGAN on images of size $256\times158$. The remarkable point is that the tunnel and car are well generated but the weather condition varies inside each image. As for the auto-encoder case, pedestrian are not easily generated.}
\end{figure}

\subsubsection{Conditional GANs}
\begin{figure}[hbt]
\centering
\includegraphics[width=0.9\linewidth]{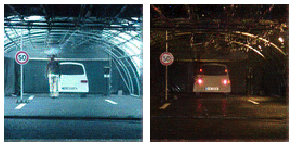}
\includegraphics[width=0.9\linewidth]{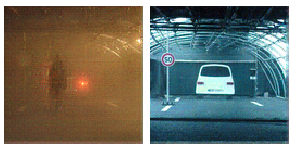}
\caption{\label{fig:cGAN} Examples of generated images using a conditional DCGAN-like architecture on images of size $128 \times 128$. The weather is uniform on each image but the pedestrian is not well generated or ignored.}
\end{figure}

For conditional GANs \cite{mirza_conditional_2014}, we use the same architecture as previously but with conditional inputs, and conditioning with the weather. As we can see on fig.\ref{fig:GAN}, in the previous generated images, several weathers can be present in a single image, but that behaviour disappears when conditioning on the weather. We illustrate that using the CEREMA database which at first sight is a very simple database with a fixed background for each weather. We can also infer that the discriminator relies at first on the background to distinguish real images from generated ones. That can back the idea that the networks learn first the most general patterns before learning the particular ones \cite{arpit_closer_2017} . As we can see in the examples on fig.\ref{fig:cGAN}, while training our conditional DCGAN, the generator learn first to generate the background for the different weather conditions properly. And only after that, we can see the generator trying to print the pedestrian on the picture. This behavior is seen for all the settings we have tried so far.

%% file: datasheet.tex
\section*{\label{sec:datasheet}Datasheet for Cerema AWP}

\noindent\color{bluep}{\framebox[\linewidth]{\bf \centering Motivation for Dataset Creation}}

\vspace{0.2cm}
\small
\noindent \color{bluep}{\bf Why was the dataset created?} (e.g., was there a specific task in mind? was there a specific gap that needed to be filled?)\color{black} \\
The Cerema AWP dataset was created for the specific need of pedestrian detection under any weather and light condition. The argument was that most of the vision techniques work pretty well when the weather is good under the day light. This dataset aimed at evaluating the existing pedestrian detectors in various conditions, and at providing a dataset to train more efficient methods.

\noindent \color{bluep}{\bf What (other) tasks could the dataset be used for?}
\color{black}\\
By construction, there are many information included in the dataset : the weather condition, the pedestrian position, the pedestrian clothes color and the pedestrian direction. It could be then used for various classification, multi-label, regression tasks. Another application is the image generation field. Indeed, the dataset is large enough to train GANs or auto-encoders. With all the available labels, this dataset could be used to study different types of conditioning in image generation or information discovering.

\noindent \color{bluep}{\bf  Has the dataset been used for any tasks already? If so, where are the results so others can compare} \color{bluep}(e.g., links to published papers)? \color{black}\\
Currently the dataset was only used in the paper presenting it in the vision field \cite{dahmane_cerema_2017}, to compare HOG and Haar detectors. 

\noindent \color{bluep}{\bf Who funded the creation of the dataset?}\color{black} \\
The Cerema organized the data collection. From the "README AND LICENCE" file available on the website: {\em This work has been sponsored by the French government research program "Investissements d'Avenir" through the IMobS3 Laboratory of Excellence (ANR-10-LABX-16-01) and the RobotEx Equipment of Excellence (ANR-10-EQPX-44), by the European Union through the Regional Competitiveness and Employment program -2014-2020- (ERDF – AURA region) and by the AURA region.
The Cerema also would like to thank Pascal Institute and the National Engineering School of Sousse.} 

\noindent \color{bluep}{\bf Any other comments?}\color{black} \\
Available at \url{https://ceremadlcfmds.wixsite.com/cerema-databases}.

\vspace{0.2cm}
\normalsize
\noindent\color{bluep}{\framebox[\linewidth]{\bf \centering Dataset Composition}}
\vspace{0.1cm}

\small
\noindent \color{bluep}{\bf What are the instances?}\color{bluep} (that is, examples; e.g., documents, images, people, countries) Are there multiple types of instances? (e.g., movies, users, ratings; people, interactions between them; nodes, edges) \color{black}\\
Instances are images produced in {\em Fog and Rain R\&D Platform}. This platform is used for : reproduction and control of fog’s particle size, of meteorological visibility, and of rain’s particle size and intensity; physical characterization of natural and artificial fog and rain. The platform is used for various applications as development of algorithms for image analysis and processing in adverse weather conditions, application and development of new imaging technologies in the specific conditions of fog and rain (infrared, laser…), performance testing of upcoming advance driver assistance systems (ADAS) including automatic obstacle detection features (lidar).
\footnote{More information - in French - \url{https://www.cerema.fr/system/files/documents/technology2017/Cerema-effi-sciences_Plateforme-9-10.pdf}}. 

\noindent \color{bluep}{\bf Are relationships between instances made explicit in the data?} (e.g., social network links, user/movie ratings, etc.)?\color{black} No\\
\noindent \color{bluep}{\bf How many instances are there? (of each type, if appropriate)?}\color{black}\\
There are 51302 images.

\begin{figure}[ht]
\includegraphics[width=\linewidth]{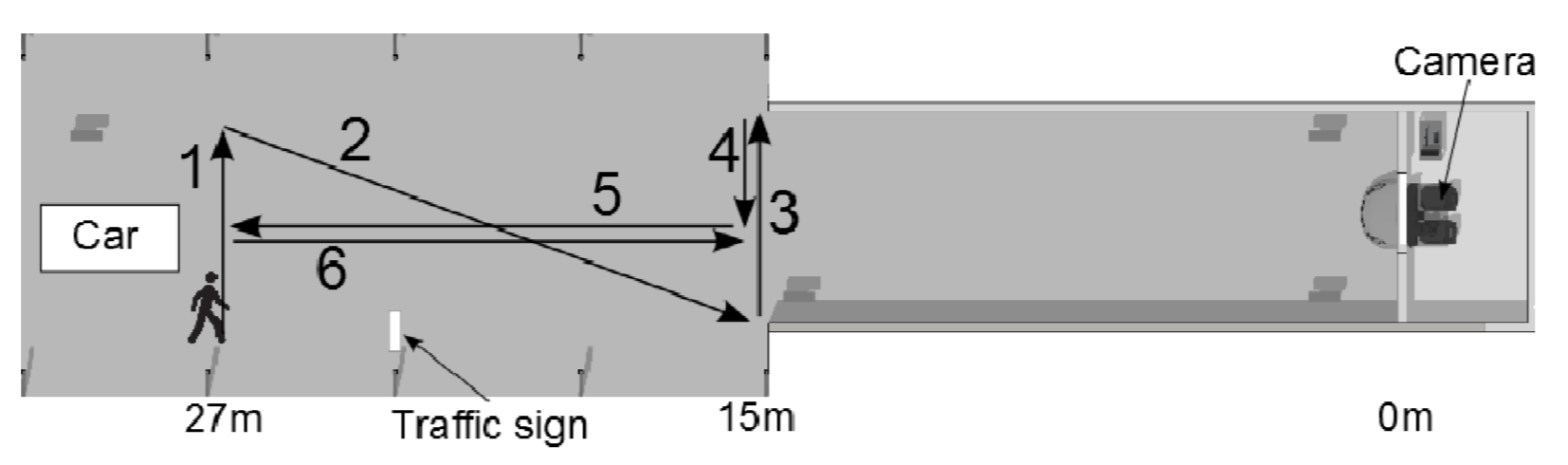}
\caption{\label{fig:tunnel} Path taken by each of the pedestrian, indicating the directions.}
\end{figure}

\noindent \color{bluep}{\bf What data does each instance consist of? “Raw” data} (e.g., unprocessed text or images)? Features/attributes? Is there a label/target associated with instances? If the instances related to people, are subpopulations identified (e.g., by age, gender, etc.) and what is their distribution?\color{black}\\
Images are raw images, in resolution 1024x631, which corresponds to standard resolution for cameras in cars.  
Concerning the image descriptions, some are available according to the folder organization (folder names give the classes)
\begin{itemize}
	\item the weather : 10 classes, day and night, each with good weather, light rain, heavy rain, light fog, heavy fog
    \item the pedestrian : 5 groups, corresponding to 5 people who participated to the data collection : 3 women, 2 men. 
    \item the clothes : 2 categories, each pedestrian appears with 2 outfits (one light and one dark)
\end{itemize}
Other labels are available in an external csv file:
\begin{itemize}
	\item the bounding box containing the pedestrian (coordinates of the top left corner, height and width)
    \item the direction of the pedestrian : 6 directions, left close, right far and right close, forth, back and 1 diagonal. See figure \ref{fig:tunnel}
    \item pedestrian id : 10 groups, corresponding to each of the 5 persons with each of the 2 outfits. Be careful, some labels are set to 0 in the file (but it is easy to correct it from the pedestrian and clothes information. 
\end{itemize}

\noindent \color{bluep}{\bf Is everything included or does the data rely on external resources?} (e.g., websites, tweets, datasets) If external resources, a) are there guarantees that they will exist, and remain constant, over time; b) is there an official archival version; c) are there access restrictions or fees?\color{black}\\
All included

\noindent \color{bluep}{\bf Are there recommended data splits and evaluation measures?} (e.g., training, development, testing; accuracy or AUC)\color{black} No

\noindent \color{bluep}{\bf  What experiments were initially run on this dataset?} \color{black}\\
Pedestrian detection was tested only in favorable conditions. So that,  Cerema-AWP was created to evaluate pedestrian detection under adverse weather conditions by day and by night with two of the most common detectors (Haar detector \cite{viola_rapid_2001} and HOG detector \cite{dalal_histograms_2005})

\noindent \color{bluep}{\bf  Have a summary of those results.} \color{black}\\
The results show that two representative state-of-the-art detectors (HOG and Haar) have much lower results when the weather is degraded, which justifies the need for this dataset. 

\noindent \color{bluep}{\bf Any other comments?}\color{black}\\
If necessary, additional experiments can be carried out within the Fog and Rain R\&D Platform on request. For this, it is possible to contact the authors (Pierre Duthon)

\vspace{0.2cm}
\normalsize
\noindent\color{bluep}{\framebox[\linewidth]{\bf \centering Data Collection Process}}
\vspace{0.05cm}

\small
\noindent \color{bluep}{\bf  How was the data collected?} (e.g., hardware apparatus/sensor, manual human curation, software program, software interface/API) \color{black}\\
The images were acquired by DFW-SX700 Sony camera at 7.5 fps and with a resolution of 1024x631 (aperture = 8.6 ; focus = 30m ; zoom = 8mm ; exposure time given for each image in the csv file).

\noindent \color{bluep}{\bf Who was involved in the data collection process?}
(e.g., students, crowdworkers) and how were they compensated (e.g., how much were crowdworkers paid)?\color{black}\\
3 permanent researchers, 2 permanent technicians and 1 trainee was 
involved for the technical preparation of data collection. 1 permanent 
technician, 1 PhD student, 1 postdoc and 2 trainees are pedestrians on 
the dataset. Labeling was performed by the permanent researcher and a 
trainee.

\noindent \color{bluep}{\bf Over what time-frame was the data collected? Does the collection time-frame match the creation time-frame of the instances?}\color{black}\\
It was collected between April 19th and 27th 2016. The real time-frame of the instance is given in the filename of each instance in format YYYY-MM-DD\_HH-MM-SS\_MS.

\noindent \color{bluep}{\bf  How was the data associated with each instance acquired?} Was the data directly observable (e.g., raw text, movie ratings), reported by subjects (e.g., survey responses), or indirectly inferred/derived from other data (e.g., part of speech tags; model-based guesses for age or language)? If the latter two, were they validated/verified and if so how?\color{black}\\
All information are directly observable and the collection process was done according to a scenario. There's no room for subjectivity.

\noindent \color{bluep}{\bf  Does the dataset contain all possible instances? Or is it a sample (not necessarily random) of instances from a larger set?}\color{black}
It contains all possible instances (all positions of the scenario for all pedestrian under all weather conditions).

\noindent \color{bluep}{\bf If the dataset is a sample, then what is the population? What was the sampling strategy} (e.g., deterministic, probabilistic with specific sampling probabilities)? Is the sample representative of the larger set (e.g., geographic coverage)? If not, why not (e.g., to cover a more diverse range of instances)? How does this affect possible uses?\color{black}\\
Not relevant.

\noindent \color{bluep}{\bf  Is there information missing from the dataset and why?} (this does not include intentionally dropped in- stances; it might include, e.g., redacted text, withheld documents) Is this data missing because it was unavailable?\color{black}
No.

\noindent \color{bluep}{\bf Any other comments?}

\vspace{0.2cm}
\normalsize
\noindent\color{bluep}{\framebox[\linewidth]{\bf \centering Preprocessing}}
\vspace{0.05cm}

\small
\noindent \color{bluep}{\bf  What preprocessing/cleaning was done?} (e.g., discretization or bucketing, tokenization, part-of-speech tagging, SIFT feature extraction, removal of instances)
Was the “raw” data saved in addition to the preprocessed/cleaned data? (e.g., to support unanticipated future uses)\color{black}\\
Only raw data.

\noindent \color{bluep}{\bf Is the preprocessing software available?}\color{black}\\
Not relevant.

\noindent \color{bluep}{\bf Does this dataset collection/processing procedure achieve the motivation for creating the dataset stated in the first section of this datasheet? If not, what are the limitations?}\color{black}
Yes

\noindent \color{bluep}{\bf Any other comments?}\color{black}

\vspace{0.2cm}
\normalsize
\noindent\color{bluep}{\framebox[\linewidth]{\bf \centering Dataset Distribution}}
\vspace{0.05cm}

\small
\noindent \color{bluep}{\bf How will the dataset be distributed?} (e.g., tarball on website, API, GitHub; does the data have a DOI and is it archived redundantly?)\color{black}\\
3 rar files on the website.

\noindent \color{bluep}{\bf When will the dataset be released/first distributed? What license (if any) is it distributed under?}\color{black}\\
It was released in February 2018 under the Open Database Licence  \url{http://opendatacommons.org/licenses/odbl/1.0/}.

\noindent \color{bluep}{\bf  Are there any copyrights on the data?}\color{black}\\
Any rights in individual contents of the database are licensed under the Database Contents License: {\url http://opendatacommons.org/licenses/dbcl/1.0/}.

\noindent \color{bluep}{\bf Are there any fees or access/export restrictions?}\color{black}
No

\noindent \color{bluep}{\bf Any other comments?}\color{black}

\vspace{0.2cm}
\normalsize
\noindent\color{bluep}{\framebox[\linewidth]{\bf \centering Dataset Maintenance}}
\vspace{0.05cm}

\small
\noindent \color{bluep}{\bf Who is supporting/hosting/maintaining the dataset?}\color{black}\\
Cerema.

\noindent \color{bluep}{\bf Will the dataset be updated? If so, how often and by whom?}\color{black}
No.

\noindent \color{bluep}{\bf How will updates be communicated?} (e.g., mailing list, GitHub)\color{black}\\
Not relevant.

\noindent \color{bluep}{\bf Is there an erratum?} \color{black}
No.

\noindent \color{bluep}{\bf If the dataset becomes obsolete how will this be communicated?}\color{black}\\
Not relevant.

\noindent \color{bluep}{\bf Is there a repository to link to any/all papers/systems that use this dataset?}\color{black}
No.

\noindent \color{bluep}{\bf If others want to extend/augment/build on this dataset, is there a mechanism for them to do so? If so, is there a process for tracking/assessing the quality of those contributions. What is the process for communicating/distributing these contributions to users?}\color{black}
No.

\noindent \color{bluep}{\bf Any other comments?}\color{black}

\vspace{0.2cm}
\normalsize
\noindent\color{bluep}{\framebox[\linewidth]{\bf \centering Legal \& Ethical Considerations}}
\vspace{0.05cm}

\small
\noindent \color{bluep}{\bf If the dataset relates to people (e.g., their attributes) or was generated by people, were they informed about the data collection?} (e.g., datasets that collect writing, photos, interactions, transactions, etc.)\color{black}
Not relevant.

\noindent \color{bluep}{\bf If it relates to people, were they told what the dataset would be used for and did they consent?} If so, how? Were they provided with any mechanism to revoke their consent in the future or for certain uses?\color{black}
Not relevant.

\noindent \color{bluep}{\bf If it relates to people, could this dataset expose people to harm or legal action?} (e.g., financial social or otherwise) What was done to mitigate or reduce the potential for harm?\color{black}
Not relevant.

\noindent \color{bluep}{\bf If it relates to people, does it unfairly advantage or disadvantage a particular social group? In what ways? How was this mitigated?} \color{black}
Not relevant.

\noindent \color{bluep}{\bf If it relates to people, were they provided with privacy guarantees? If so, what guarantees and how are these ensured?}\color{black}
Not relevant.

\noindent \color{bluep}{\bf Does the dataset comply with the EU General Data Protection Regulation (GDPR)? Does it comply with any other standards, such as the US Equal Employment Opportunity Act?}\color{black}
Not relevant.

\noindent \color{bluep}{\bf  Does the dataset contain information that might be considered sensitive or confidential?} (e.g., personally identifying information)\color{black}
No.

\noindent \color{bluep}{\bf  Does the dataset contain information that might be considered inappropriate or offensive?} \color{black} 
No.

\noindent \color{bluep}{\bf Any other comments?}
\color{black}
\normalsize

%% file: appendix.tex
\appendix{}

\section{Architecture details}
\subsection{Classification and regression tasks}
The provided architecture is for Weather and PedestrianClothes which are both 10 classes. For the other classification tasks, only the output is modified to match the number of classes or the number of regression values.
\scriptsize
\begin{verbatim}
Layer (type)                 Output Shape              Param #   
=================================================================
conv1 (Conv2D)               (None, 77, 126, 64)       4864      
_________________________________________________________________
pool1 (MaxPooling2D)         (None, 38, 63, 64)        0         
_________________________________________________________________
batch_normalization_1 (Batch (None, 38, 63, 64)        256       
_________________________________________________________________
conv2 (Conv2D)               (None, 18, 31, 128)       73856     
_________________________________________________________________
pool2 (MaxPooling2D)         (None, 9, 15, 128)        0         
_________________________________________________________________
batch_normalization_2 (Batch (None, 9, 15, 128)        512       
_________________________________________________________________
conv3 (Conv2D)               (None, 4, 7, 256)         295168    
_________________________________________________________________
pool3 (MaxPooling2D)         (None, 2, 3, 256)         0         
_________________________________________________________________
batch_normalization_3 (Batch (None, 2, 3, 256)         1024      
_________________________________________________________________
flatten_19 (Flatten)         (None, 1536)              0         
_________________________________________________________________
fc2 (Dense)                  (None, 128)               196736    
_________________________________________________________________
output_weather (Dense)       (None, 10)                1290      
=================================================================
\end{verbatim}

\subsection{Multi label classification}
\scriptsize
\begin{verbatim}
Layer (type)         Output Shape         Param#  Connected to                     
================================================================
main_input(InputLayer) (None, 158, 256, 3)  0
________________________________________________________________
conv1 (Conv2D)       (None, 77, 126, 64)  4864 main_input[0][0]
________________________________________________________________
pool1 (MaxPooling2D)   (None, 38, 63, 64)   0      conv1[0][0]
________________________________________________________________
batch_norm_13(BatchNorm)(None, 38, 63, 64)   256   pool1[0][0]
________________________________________________________________
conv2 (Conv2D)  (None, 18, 31, 128)  73856   batch_norm_13[0][0]
________________________________________________________________
pool2 (MaxPooling2D)   (None, 9, 15, 128)   0      conv2[0][0]
________________________________________________________________
batch_norm_14(BatchNorm) (None, 9, 15, 128)   512  pool2[0][0]
________________________________________________________________
conv3 (Conv2D)    (None, 4, 7, 256)  295168  batch_norm_14[0][0]    
________________________________________________________________
pool3 (MaxPooling2D)   (None, 2, 3, 256)    0       conv3[0][0]
________________________________________________________________
batch_norm_15 BatchNorm) (None, 2, 3, 256)  1024    pool3[0][0] 
________________________________________________________________
flatten_5 (Flatten)   (None, 1536)    0      batch_norm_15[0][0]
________________________________________________________________
fc1d (Dense)           (None, 512)       786944  flatten_5[0][0]
________________________________________________________________
fc1w (Dense)           (None, 512)       786944  flatten_5[0][0]
________________________________________________________________
fc1p (Dense)           (None, 512)       786944  flatten_5[0][0]
________________________________________________________________
fc2d (Dense)           (None, 128)       65664   fc1d[0][0]
________________________________________________________________
fc2w (Dense)           (None, 128)       65664   fc1w[0][0]
________________________________________________________________
fc2p (Dense)           (None, 128)       65664   fc1p[0][0]
________________________________________________________________
output_direction(Dense) (None, 6)        774     fc2d[0][0]
________________________________________________________________
output_weather (Dense) (None, 6)         774     fc2w[0][0]
________________________________________________________________
output_pedestrian (Dense) (None, 5)      645     fc2p[0][0]
________________________________________________________________
output_clothes (Dense)  (None, 2)        258     fc2p[0][0]
================================================================
\end{verbatim}

\subsection{Auto-encoder}
\scriptsize
\begin{verbatim}
_________________________________________________________________
Layer (type)                 Output Shape              Param #   
=================================================================
image_input (InputLayer)     (None, 158, 256, 3)       0         
_________________________________________________________________
conv1 (Conv2D)               (None, 79, 128, 64)       4864      
_________________________________________________________________
pool1 (MaxPooling2D)         (None, 40, 64, 64)        0         
_________________________________________________________________
batch_normalization_85 (Batc (None, 40, 64, 64)        256       
_________________________________________________________________
conv2 (Conv2D)               (None, 20, 32, 128)       73856     
_________________________________________________________________
pool2 (MaxPooling2D)         (None, 10, 16, 128)       0         
_________________________________________________________________
batch_normalization_86 (Batc (None, 10, 16, 128)       512       
_________________________________________________________________
conv3 (Conv2D)               (None, 5, 8, 256)         295168    
_________________________________________________________________
pool3 (MaxPooling2D)         (None, 3, 4, 256)         0         
_________________________________________________________________
batch_normalization_87 (Batc (None, 3, 4, 256)         1024      
_________________________________________________________________
flatten_29 (Flatten)         (None, 3072)              0         
_________________________________________________________________
fc1d (Dense)                 (None, 256)               786688    
_________________________________________________________________
fc2d (Dense)                 (None, 100)               25700     
_________________________________________________________________
fc1g (Dense)                 (None, 256)               25856     
_________________________________________________________________
fc2g (Dense)                 (None, 3072)              789504    
_________________________________________________________________
reshape_27 (Reshape)         (None, 3, 4, 256)         0         
_________________________________________________________________
deconv3 (Conv2DTranspose)    (None, 6, 8, 256)         590080    
_________________________________________________________________
cropping2d_15 (Cropping2D)   (None, 5, 8, 256)         0         
_________________________________________________________________
depool2 (UpSampling2D)       (None, 10, 16, 256)       0         
_________________________________________________________________
deconv2 (Conv2DTranspose)    (None, 20, 32, 128)       295040    
_________________________________________________________________
depool1 (UpSampling2D)       (None, 40, 64, 128)       0         
_________________________________________________________________
deconv1 (Conv2DTranspose)    (None, 80, 128, 64)       204864    
_________________________________________________________________
depool0 (UpSampling2D)       (None, 160, 256, 64)      0         
_________________________________________________________________
cropping2d_16 (Cropping2D)   (None, 158, 256, 64)      0         
_________________________________________________________________
deconv0 (Conv2DTranspose)    (None, 158, 256, 3)       4803      
=================================================================
\end{verbatim}

\subsection{GAN}
\begin{verbatim}
Generator:
_________________________________________________________________
Layer (type)                 Output Shape              Param #   
=================================================================
dense_3 (Dense)              (None, 81920)             4177920   
_________________________________________________________________
leaky_re_lu_8 (LeakyReLU)    (None, 81920)             0         
_________________________________________________________________
batch_normalization_5 (Batch (None, 81920)             327680    
_________________________________________________________________
reshape_2 (Reshape)          (None, 20, 32, 128)       0         
_________________________________________________________________
dropout_5 (Dropout)          (None, 20, 32, 128)       0         
_________________________________________________________________
up_sampling2d_4 (UpSampling2 (None, 40, 64, 128)       0         
_________________________________________________________________
conv2d_transpose_5 (Conv2DTr (None, 40, 64, 128)       409728    
_________________________________________________________________
leaky_re_lu_9 (LeakyReLU)    (None, 40, 64, 128)       0         
_________________________________________________________________
batch_normalization_6 (Batch (None, 40, 64, 128)       512       
_________________________________________________________________
up_sampling2d_5 (UpSampling2 (None, 80, 128, 128)      0         
_________________________________________________________________
conv2d_transpose_6 (Conv2DTr (None, 80, 128, 64)       204864    
_________________________________________________________________
cropping2d_2 (Cropping2D)    (None, 79, 128, 64)       0         
_________________________________________________________________
leaky_re_lu_10 (LeakyReLU)   (None, 79, 128, 64)       0         
_________________________________________________________________
batch_normalization_7 (Batch (None, 79, 128, 64)       256       
_________________________________________________________________
up_sampling2d_6 (UpSampling2 (None, 158, 256, 64)      0         
_________________________________________________________________
conv2d_transpose_7 (Conv2DTr (None, 158, 256, 32)      51232     
_________________________________________________________________
leaky_re_lu_11 (LeakyReLU)   (None, 158, 256, 32)      0         
_________________________________________________________________
batch_normalization_8 (Batch (None, 158, 256, 32)      128       
_________________________________________________________________
conv2d_transpose_8 (Conv2DTr (None, 158, 256, 3)       2403      
_________________________________________________________________
activation_3 (Activation)    (None, 158, 256, 3)       0         
=================================================================
Total params: 5,174,723
Trainable params: 5,010,435
Non-trainable params: 164,288
_________________________________________________________________
Discriminator:
_________________________________________________________________
Layer (type)                 Output Shape              Param #   
=================================================================
conv2d_4 (Conv2D)            (None, 79, 128, 32)       2432      
_________________________________________________________________
leaky_re_lu_12 (LeakyReLU)   (None, 79, 128, 32)       0         
_________________________________________________________________
dropout_6 (Dropout)          (None, 79, 128, 32)       0         
_________________________________________________________________
conv2d_5 (Conv2D)            (None, 40, 64, 64)        51264     
_________________________________________________________________
leaky_re_lu_13 (LeakyReLU)   (None, 40, 64, 64)        0         
_________________________________________________________________
dropout_7 (Dropout)          (None, 40, 64, 64)        0         
_________________________________________________________________
conv2d_6 (Conv2D)            (None, 20, 32, 128)       204928    
_________________________________________________________________
leaky_re_lu_14 (LeakyReLU)   (None, 20, 32, 128)       0         
_________________________________________________________________
dropout_8 (Dropout)          (None, 20, 32, 128)       0         
_________________________________________________________________
flatten_2 (Flatten)          (None, 81920)             0         
_________________________________________________________________
dense_4 (Dense)              (None, 1)                 81921     
_________________________________________________________________
activation_4 (Activation)    (None, 1)                 0         
=================================================================
Total params: 340,545
Trainable params: 340,545
Non-trainable params: 0
_________________________________________________________________
\end{verbatim}